%% file: 0_main.tex
\documentclass{cambrian}
\usepackage[utf8]{inputenc} 
\usepackage[T1]{fontenc}    
\usepackage{hyperref}       
\usepackage{url}            
\usepackage{booktabs}       
\usepackage{amsfonts}       
\usepackage{nicefrac}       
\usepackage{microtype}      

\usepackage{mwe}
\usepackage{graphicx}
\usepackage{amssymb}
\usepackage{tabularray}
\usepackage{adjustbox}
\usepackage{colortbl}
\usepackage{array}
\usepackage{multirow}
\usepackage{makecell}
\usepackage{bbm}
\usepackage{collcell,xfp}
\usepackage{pgf}
\usepackage{tikz}
\usepackage[most]{tcolorbox}
\usepackage{csquotes}
\usepackage[noorphans,vskip=1em,leftmargin=1em]{quoting}
\usepackage{pifont} 
\usepackage{enumitem} 
\usepackage{tcolorbox} 
\usepackage{forest}
\usepackage{wrapfig}

\usepackage{caption}
\usepackage{subcaption}
\usepackage{longtable}
\usepackage{algorithm}
\usepackage{algorithmic}
\usepackage{amsmath}
\usepackage[capitalize]{cleveref}
\crefname{algorithm}{Algorithm}{Algorithms}
\Crefname{algorithm}{Algorithm}{Algorithms}
\usepackage[percent]{overpic}
\usepackage{xspace}
\usepackage[normalem]{ulem}  
\usepackage{tocloft}  
\usepackage[numbers]{natbib}
\usepackage{mathtools}
\usepackage{listings}

\usepackage{enumitem} 
\usepackage{multirow}
\usepackage{amssymb}

\usepackage{bm}
\usepackage{bbm}
\usepackage{soul}
\usepackage{fontawesome}

\newcommand{\github}{\raisebox{-1.5pt}{\includegraphics[height=1.05em]{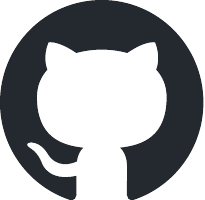}}\xspace}
\newcommand{\worldwideweb}{\raisebox{-1.5pt}{\includegraphics[height=1.05em]{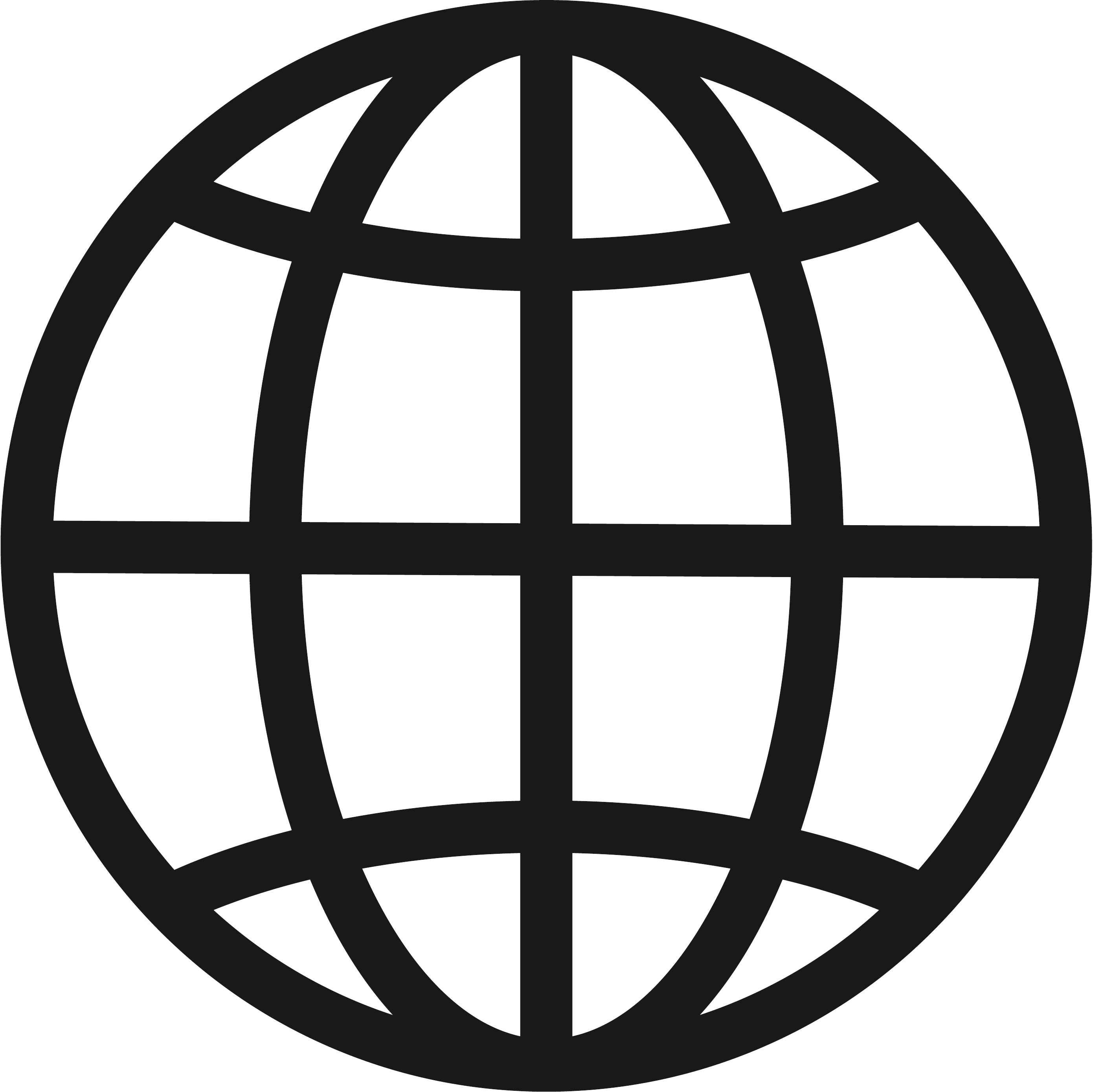}}\xspace}

\definecolor{custom_navy}{RGB}{30, 65, 155}
\definecolor{gg}{gray}{0.92}
\newcolumntype{a}{>{\columncolor{gg}}c}

\definecolor{scholarblue}{rgb}{0.21,0.49,0.74}
\definecolor{bluelink}{RGB}{0,113,188}
\definecolor{greenlink}{RGB}{0,188,113}
\hypersetup{
    colorlinks=true,%
    citecolor=scholarblue,%
    filecolor=red,%
    linkcolor=red!93!black,%
    urlcolor=bluelink
}

\usepackage{titlesec}
\titlespacing*{\paragraph}
    {0pt}     
    {0.25em}  
    {1em}  

\definecolor{navyblue}{HTML}{0071BC}
\iftrue   
\newcommand{\displaytodo}[1]{#1}
\else
\newcommand{\displaytodo}[1]{}
\fi

\definecolor{diffgreen}{RGB}{213, 245, 213}
\definecolor{diffred}{RGB}{255, 215, 215}

\definecolor{diffgreen}{RGB}{213, 245, 213}
\definecolor{diffred}{RGB}{255, 215, 215}

\definecolor{blindcolor}{HTML}{AB2AC6}    
\definecolor{chancecolor}{HTML}{F59E0B}   
\definecolor{singlecolor}{HTML}{06B6D4}   
\definecolor{multiplecolor}{HTML}{2563EB} 
\definecolor{captioncolor}{HTML}{22C55E}  

 \newcommand{\culine}[2]{%
    \def\temp@uline{\bgroup\markoverwith
        {\textcolor{#1}{\rule[-0.5ex]{2pt}{1pt}}}\ULon}%
    \temp@uline{#2}%
}
 \newcommand{\cthickuline}[3][0.8pt]{%
    \def\temp@uline{\bgroup\markoverwith
        {\textcolor{#2}{\rule[-0.5ex]{2pt}{#1}}}\ULon}%
    \temp@uline{#3}%
}

\newcommand{\metabbr}{\mbox{\textsc{DreamFree}}\xspace}

\title{\center{Future Representation Distillation \\ for Imagination-free World Action Models}}

\renewcommand\Affilfont{\normalfont\fontsize{11}{15}\selectfont\centering}
\author{
    Jaehyeong~Jo$^{*}$\quad
    Sangwon~Jang$^{*}$\quad
    Sihyun~Yu\quad
    Seonghyeon~Ye\quad\\
    Yekyung~Nah\quad
    Hojung~Jung\quad
    Kimin~Lee$^{\dagger}$\quad
    Sung~Ju~Hwang$^{\dagger}$\quad\\
    KAIST
}
\begingroup

\begin{NoHyper}
\footnotetext{$^*$ Equal contribution.}
\footnotetext{$^\dagger$ Equal advising.}
\end{NoHyper}
\endgroup

\input{marco}

\input{1_abstract}

\begin{document}
\maketitle


\vspace{15pt}
\begin{center}
    \renewcommand{\arraystretch}{1.5}
    \begin{tabular}{rll}
        \worldwideweb{} & \textbf{Website} &    \url{https://harryjo97.github.io/dreamfree/}\\
        \github{} & \textbf{Code} & \url{https://github.com/harryjo97/dreamfree}\\
    \end{tabular}
\end{center}


\newpage
{
    \hypersetup{linkcolor=black}
    \tableofcontents
}
\newpage

\input{2_introduction}

\input{3_related_work}
\input{4_method}
\input{5_experiments}
\input{6_conclusion}

\addcontentsline{toc}{section}{References}
\bibliography{reference}
\bibliographystyle{plainnat}

\newpage
\input{7_appendix}

\end{document}

%% file: marco.tex
\tcbuselibrary{skins, breakable} 
\definecolor{cornellred}{rgb}{0.7, 0.11, 0.11}
\definecolor{cadmiumgreen}{rgb}{0.0, 0.42, 0.24}
\definecolor{aliceblue}{rgb}{0.91, 0.94, 0.97}
\definecolor{darkblue}{rgb}{0.83, 0.89, 0.97}
\definecolor{Blue9}{rgb}{0.098,0.3,0.9}
\definecolor{mutedviolet}{HTML}{EEDAFF} 
\definecolor{mutedpink}{HTML}{FFD9D9} 
\definecolor{mutedblue}{HTML}{CCE5FF} 
\definecolor{mutedgreen}{HTML}{CEF8C2} 
\definecolor{forestgreen}{rgb}{0.13, 0.55, 0.13}
\definecolor{hookergreen}{rgb}{0.0, 0.44, 0.0}
\definecolor{cozygreen}{RGB}{170, 210, 160} 
\definecolor{vviolet}{HTML}{676aa4} 
\definecolor{vpink}{HTML}{bb7699} 
\definecolor{vblue}{HTML}{618eb4} 
\definecolor{vgreen}{HTML}{367660} 
\definecolor{vgreen}{HTML}{367660} 
\definecolor{bananamania}{rgb}{0.98, 0.91, 0.71}

\definecolor{navyblue}{HTML}{0071BC}
\definecolor{hotpink}{HTML}{FF0080}

\definecolor{oai-white}{HTML}{FFFFFF}
\definecolor{oai-black}{HTML}{000000}
\definecolor{oai-red}{HTML}{FF4500}
\definecolor{oai-green}{HTML}{51DA4C}
\definecolor{oai-blue}{HTML}{0000FF}
\definecolor{oai-yellow}{HTML}{FFF639}
\definecolor{oai-magenta}{HTML}{FF45FF}
\definecolor{oai-cyan}{HTML}{00FFFF}
\definecolor{oai-orange}{HTML}{FE7600}
\definecolor{oai-violet}{HTML}{8A2BE2}
\definecolor{oai-brown}{HTML}{A0522D}
\definecolor{oai-green-050}{HTML}{F4FFF4}
\definecolor{oai-green-100}{HTML}{E9FFE8}
\definecolor{oai-green-200}{HTML}{D9FFD8}
\definecolor{oai-green-300}{HTML}{C9FFC7}
\definecolor{oai-green-400}{HTML}{A6FFA3}
\definecolor{oai-green-500}{HTML}{7CF178}
\definecolor{oai-green-600}{HTML}{51DA4C}
\definecolor{oai-green-700}{HTML}{3FA93B}
\definecolor{oai-green-800}{HTML}{2D712A}
\definecolor{oai-green-900}{HTML}{193718}
\definecolor{oai-gray-000}{HTML}{FFFFFF}
\definecolor{oai-gray-100}{HTML}{FAFAFA}
\definecolor{oai-gray-200}{HTML}{F5F5F5}
\definecolor{oai-gray-300}{HTML}{E5E5E5}
\definecolor{oai-gray-400}{HTML}{FFB7A4}
\definecolor{oai-gray-500}{HTML}{CDCDCD}
\definecolor{oai-gray-600}{HTML}{A8A8A8}
\definecolor{oai-gray-700}{HTML}{747474}
\definecolor{oai-gray-800}{HTML}{393939}
\definecolor{oai-gray-900}{HTML}{000000}

\definecolor{tablesectionbg}{HTML}{FAF4F8} 
\definecolor{stateelementtable}{HTML}{E7CCEF}   
\definecolor{statestructuretable}{HTML}{F3C6DA} 
\definecolor{softteal}{HTML}{74C9BD}
\definecolor{softpeach}{HTML}{E99A7A}

\definecolor{statefaint}{HTML}{F0F0F0}

\definecolor{elem}{HTML}{9E35BE}
\definecolor{elembg}{HTML}{9E35BE}

\definecolor{struct}{HTML}{D9578F}
\definecolor{structbg}{HTML}{F3C6DA}

\definecolor{perception}{HTML}{5EC26A}
\definecolor{perceptionbg}{HTML}{5EC26A}

\hypersetup{
  linkcolor = cornellred,
  citecolor  = cadmiumgreen,
  colorlinks = true,
  urlcolor = Blue9
}

\newcommand*{\ShowNotes}{} 

\ifdefined\ShowNotes
  \newcommand{\colornote}[3]{{\color{#1}\bf{#2: #3}\normalfont}}
\else
  \newcommand{\colornote}[3]{}
\fi

\definecolor{ggreen}{rgb}{0.53, 0.69, 0.43}
\definecolor{gred}{HTML}{F5433D}

%% file: 1_abstract.tex
\begin{abstract}
World Action Models (WAMs) use the predictive knowledge of pretrained video diffusion models for robot control through future video prediction. However, iterative video denoising is costly, and inaccurate visual predictions can compromise action prediction. Our key idea is to make this predictive knowledge available through visual representations computed from the current observation, enabling action prediction without generating future videos. We introduce \metabbr, a WAM that repurposes a pretrained video diffusion model as a predictive visual encoder through representation distillation. The visual encoder learns to predict a frozen video diffusion model's representation using only the current observation. An action module is then trained to predict actions by extracting action-relevant information from the visual encoder. Across real-world and simulated tasks, \metabbr achieves higher success rates than state-of-the-art VLAs and WAMs, while performing real-time closed-loop control at 11Hz with a 14B video diffusion backbone. Experiments show that \metabbr can translate stronger video priors and additional video data into better robot control and enable learning new robot tasks from video demonstrations, including videos from a different embodiment, without action annotation for the new tasks.
\end{abstract}

%% file: 2_introduction.tex
\begin{figure}[t!]
    \centering
\begin{minipage}{0.29\linewidth}
    \centering
    \includegraphics[width=1\linewidth]{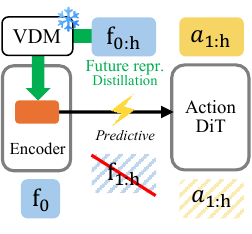}
\end{minipage}
\hfill
\begin{minipage}{0.26\linewidth}
    \centering
    \includegraphics[width=1\linewidth]{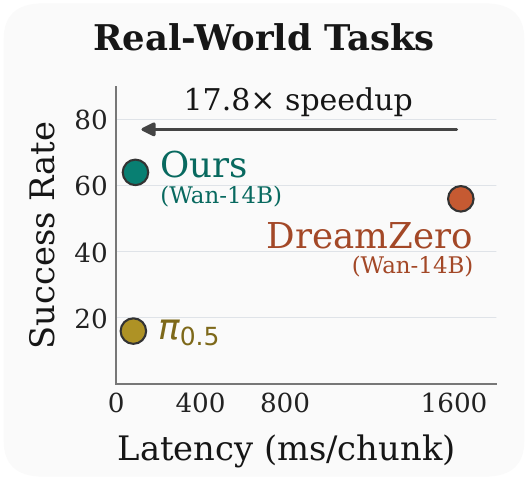}
\end{minipage}
\hfill
\begin{minipage}{0.436\linewidth}
    \centering
    \includegraphics[width=1\linewidth]{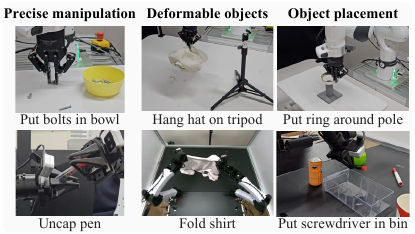}
\end{minipage}
\captionsetup{position=bottom,skip=1pt}
    \caption{\textbf{Overview}. \metabbr distills predictive knowledge from a frozen video diffusion model into a predictive visual encoder, enabling action prediction from the current observation without generating future videos. In real-world evaluations, \metabbr achieves higher success than DreamZero and $\pi_{0.5}$ with latency comparable to that of $\pi_{0.5}$ and a 17.8$\times$ inference speedup over DreamZero.
    \metabbr generalizes across diverse real-world manipulation tasks.
}\label{fig:concept}
\end{figure}

\section{Introduction}

Pretrained video diffusion models offer a promising foundation for robot control~\citep{zhu2025unified,kim2026cosmospolicy,gao2026dreamdojo}. Through large-scale video pretraining, these models learn to predict how physical environments evolve over time, providing \emph{predictive} knowledge that can guide robot actions. World Action Models (WAMs; \citet{ye2026dreamzero}) use this knowledge by adapting video models to predict visual futures and actions, demonstrating strong manipulation performance and generalization to new tasks and environments~\citep{kim2026cosmospolicy,ye2026dreamzero}.

A common approach in WAMs is to use this predictive knowledge of pretrained video diffusion models through future video prediction, either by jointly predicting visual futures and actions~\citep{kim2026cosmospolicy,ye2026dreamzero} or by decoding actions from predicted visual futures~\citep{hu2025vpp,pai2025mimicvideo}. These methods leverage the pretrained models' knowledge through their native video prediction process. However, iterative video denoising incurs substantial inference cost, and errors in the predicted visual futures can lead to inaccurate actions. Moreover, it remains unclear whether action prediction requires the same level of visual detail as video generation.

A central question is how to make better use of the predictive knowledge in pretrained video models for action prediction. The success of video-generating WAMs demonstrates the value of this knowledge, while leaving open how it should be represented to support control. This motivates investigating whether predictive information can be made directly available in the visual features used for action prediction.

We hypothesize that this predictive knowledge can be effectively captured in visual representations to support action prediction, without iterative video denoising. Our key idea is to distill the video model's predictive knowledge into the visual representations computed from the current observation. We train a visual encoder to anticipate visual futures from the observation by using the representations extracted from a pretrained video diffusion model. Our analysis shows that distillation enables the encoder to effectively capture future visual information from the current observation and make its features more useful for action prediction. The resulting predictive representations enable action prediction without generating future videos.

Based on this finding, we introduce \metabbr, a WAM that repurposes a pretrained video diffusion backbone as a predictive visual encoder through representation distillation. We train the model in two stages, separating predictive representation learning from action learning. In the first stage, a frozen video diffusion model extracts representations from both current and future frames of a video demonstration, and the visual encoder learns to predict these representations from the current observation. The encoder thus learns to anticipate scene evolution in feature space, making the video model's predictive knowledge available in a single forward pass without future video generation. This stage requires no action labels, allowing the visual encoder to learn from any in-the-wild video demonstrations. In the second stage, we train an action module on robot demonstrations to map these visual features to executable actions, with the visual encoder either frozen or jointly fine-tuned.

We evaluate \metabbr on diverse manipulation tasks in both simulation and real-world environments~\citep{khazatsky2024droid}, demonstrating stronger performance compared to state-of-the-art VLA and WAM baselines~\citep{intelligence2025pi_05,ye2026dreamzero}. 
In real-world evaluations, \metabbr succeeds on tasks involving object deformation or successive rearrangements where video-generating WAM fails (Figure~\ref{fig:concept}). 
These successes suggest a benefit of using predictive visual representations for control without relying on potentially inaccurate video predictions. Experiments show that using a larger video backbone and additional pretraining video data improve success, demonstrating \metabbr can translate stronger video priors and additional video data into better robot control without additional action annotation. We additionally show task transfer from video demonstrations, including videos from a different embodiment, showing that \metabbr can learn new tasks from videos without collecting action annotations for those tasks.

In addition to strong performance, \metabbr offers a practical advantage for real-time control. During deployment, the policy obtains predictive visual representations in a single encoder pass and uses them to generate an action chunk without iterative video denoising. \metabbr using a 14B video backbone achieves real-time closed-loop control at 11Hz, achieving a $17.8\times$ inference speedup over a video-generating WAM with the same video backbone size and latency comparable to that of a VLA (Figure~\ref{fig:concept}, middle).

We highlight our main contributions below:
\vspace{-0.05in}
\begin{itemize}[itemsep=0.5mm, parsep=3pt, leftmargin=20pt]
\item We introduce \metabbr, a WAM that repurposes a pretrained video diffusion backbone as a predictive visual encoder by distilling future information from a frozen video diffusion model.
\item We outperform state-of-the-art VLA and WAM baselines in real-world manipulation without task-specific adaptation, while achieving real-time closed-loop control at 11Hz.
\item We show that \metabbr translates stronger video priors and additional video data into better robot control.
\item We demonstrate that \metabbr can learn new tasks from video demonstrations, including videos from a different embodiment, without action annotations for the new tasks. 
\end{itemize}

%% file: 3_related_work.tex
\section{Related work}

\paragraph{Vision-language-action policies}
Vision-Language-Action (VLA) policies adapt pretrained vision-language models to map visual observations and language instructions directly to robot actions~\citep{intelligence2025pi_05, rss2025_pi0,zitkovich2023rt,bjorck2025gr00t,kim2024openvla}. While their backbones provide semantic and linguistic priors, pretraining on static image-text pairs gives them limited knowledge of the spatiotemporal dynamics important for robot control.

\paragraph{Video model-based robot policies}
Video model-based robot policies build on pretrained video diffusion models by coupling future visual prediction with action prediction. Some decode actions from video representations through inverse dynamics~\citep{hu2025vpp,pai2025mimicvideo,rhoda2026dva}, while others jointly denoise future video and action sequences~\citep{kim2026cosmospolicy,li2026lingbotva,bi2026motus,ma2026dit4dit,ye2026dreamzero}. These World Action Models (WAMs) transfer spatiotemporal video priors to control, achieving strong action-data efficiency and generalization. However, generating future video at deployment introduces substantial inference cost, and errors in the predicted video can lead to inaccurate action prediction. Recent works retain video prediction during training while enabling action prediction without future video generation at deployment~\citep{yuan2026fastwam,ye2026gigaworld}, yet how faithfully their representations preserve the video model's predictive knowledge remains unclear.

\paragraph{Predictive visual representation learning}
Predictive visual learning methods train encoders to capture scene structure and dynamics by predicting representations of unobserved visual content~\citep{assran2025vjepa}. Related predictive supervision is used in VLA policies, where future images or visual features serve as auxiliary training targets alongside action supervision~\citep{wu2024gr1,zhao2025cot,zhang2026dreamvla}. For example, FLARE~\citep{zheng2025flare} aligns future tokens in the action expert with vision-language embeddings of future observations. Our approach learns predictive visual representations from the current observation by distilling features from a frozen video diffusion model into a visual encoder.

%% file: 4_method.tex
\section{\metabbr}

\begin{figure}[t!]
    \centering
    \includegraphics[width=1\linewidth, ]{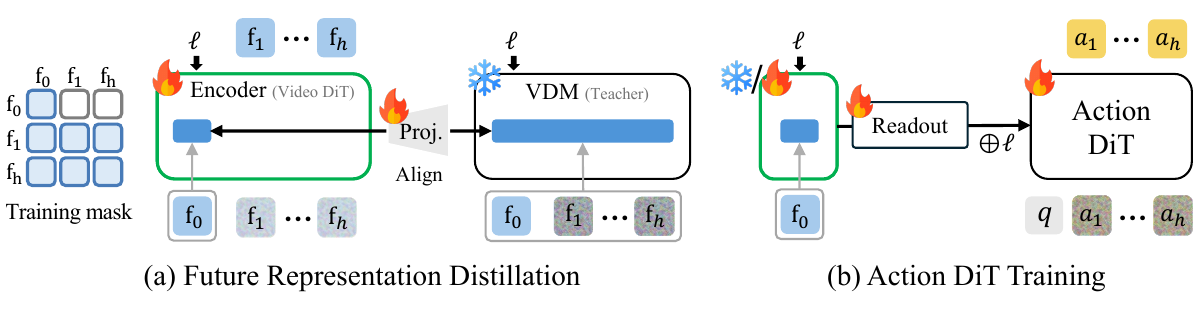}
    \captionsetup{position=bottom,skip=-1pt}
    \caption{
    \textbf{Model architecture of \metabbr}. Training is separated into two stages: (a) In the first stage, a pretrained video diffusion backbone is repurposed as a visual encoder through video generative training with a structured attention mask and distillation of predictive representations from the frozen video diffusion model. 
    (b) In the second stage, the action readout module and action decoder are trained to predict actions by extracting action-relevant information from the encoder.
    }
    \label{fig:method}
\end{figure}

As shown in Figure~\ref{fig:method}, \metabbr consists of a predictive visual encoder followed by an action module. The training stage of \metabbr is separated into two stages. First, the predictive visual encoder is trained to predict the representations from the frozen video diffusion model (Section \ref{subsection:frd}). The frozen video diffusion model extracts representations from both current and future frames of a video demonstration, while the visual encoder conditions on the current observation only.
Second, the action module is trained to predict actions by extracting action-relevant information from the predictive visual encoder through a readout module (Section \ref{subsection:action_gen}). During inference, only a single forward pass through the predictive visual encoder is required to extract visual features relevant for action prediction, enabling real-time closed-loop control up to 11Hz.

\subsection{Learning a predictive visual encoder}
\label{subsection:frd}

\paragraph{Predictive visual encoder}
We initialize the encoder from a pretrained video diffusion backbone and use the hidden states of its current-frame latent tokens as the visual representation. We apply a structured mask to the backbone's bidirectional video self-attention (Figure~\ref{fig:method}(a) left), restricting current-frame tokens to attend only to each other within self-attention. This ensures that the current-frame hidden states depend only on the current observation and instruction. During training, the backbone receives a clean current frame and noisy future frames, as shown in Figure~\ref{fig:method}(a).

\paragraph{Future representation distillation}
The main idea is to distill the representation of a frozen video diffusion model into the visual encoder that observes only the current observation and instruction.
Motivated by REPA~\citep{yu2025repa}, we introduce \emph{future representation distillation}, which aligns the projection of the encoder's current frame representation with a video diffusion model representation as shown in Figure~\ref{fig:method}(a).


Formally, let $\bm{u}_{\theta}$ be the predictive visual encoder, $\bm{g}$ be the pretrained video diffusion model, and $\bm{p}_{\psi}$ be the trainable projector. Consider a clean video $\bm{x}_{\ast}$ with the first frame $\bm{x}_{\ast}^{0}$ and language instruction $\boldsymbol{\ell}$, and let $\bm{h}_i$ be the $i$-th layer hidden representation of $\bm{u}_{\theta}(\bm{x}_{\ast}^{0},\boldsymbol{\ell})$.
We use $\bm{g}$ as the teacher and take its hidden representation at layer $j$, computed from a perturbed latent of $\bm{x}_{\ast}$, as the distillation target.
We use a lightweight Transformer decoder for the projector $\bm{p}_{\psi}$ that maps the current-frame representation $\bm{h}_i$ to the spatiotemporal token layout and feature dimension of the teacher's full-video representation.
Note that $\bm{g}$ and $\bm{p}_{\psi}$ are used only during training and discarded at deployment.

Future representation distillation is achieved by maximizing the similarity between the current-only encoder representation and the teacher representation as follows:
\begin{equation}
    \mathcal{L}_{\text{distill}}(\theta,\psi) \coloneqq \mathbb{E}_{\bm{x}_{\ast},\bm{\epsilon}_d} \big[ -\mathrm{sim}\big( \bm{p}_{\psi}(\bm{h}_i), \bm{g}_j(\bm{x}_{t_d},t_d,\boldsymbol{\ell}) \big) \big],
\end{equation}
where $\bm{\epsilon}_d$ is Gaussian noise, $t_d\in[0,1]$ is a fixed timestep controlling the teacher input’s noise level, $\bm{x}_{t_d}=(1 - t_d)\bm{x}_{\ast} + t_d \bm{\epsilon}_d$, and $\mathrm{sim}(\cdot,\cdot)$ is a pre-defined similarity function.

To preserve the predictive capability inherited from the pretrained video backbone, we also use the original generative flow matching objective as a regularization:
\begin{equation}
    \mathcal{L}_{\text{gen}}(\theta) \coloneqq \mathbb{E}_{\bm{x}_{\ast},\bm{\epsilon},t}\Big[\big\| \bm{u}_{\theta}(\bm{x}_t, t, \boldsymbol{\ell}) - (\bm{\epsilon} - \bm{x}_{\ast}) \big\|_2^2\Big],
\end{equation}
where $\bm{u}_{\theta}$ is conditioned on clean current-frame latent tokens with timestep $0$, while the future-frame latent tokens in $x_t$ are perturbed with Gaussian noise $\bm{\epsilon}$ at timestep $t$.

The encoder is trained with both the native generative loss and the distillation loss as follows:
\begin{equation}
    \mathcal{L}(\theta,\psi) \coloneqq \mathcal{L}_{\text{gen}}(\theta) + \mathcal{L}_{\text{distill}}(\theta,\psi) .
\end{equation}

\subsection{Learning actions from visual representations}
\label{subsection:action_gen}

After training the predictive visual encoder, we use action-labeled robot demonstrations to train the action module, which consists of a readout module, state and action encoders, and an action decoder. 
Given the observation and instruction, the visual encoder produces the predictive representation that serves as a visual prior for the action module. Then the readout module extracts action-relevant features from the representation, and the action module conditions on the action-relevant features together with the instruction and proprioception to predict the action chunk, as illustrated in Figure~\ref{fig:method}(b).

\paragraph{Action readout module}
Given observation $\bm{o}$ and instruction $\boldsymbol{\ell}$, we extract the hidden representation $\bm{h}_k$ from the visual encoder, where $k$ denotes the layer used for action readout. We transform these visual tokens into a fixed number of action-facing tokens using a latent-query readout inspired by Perceiver~\citep{jaegle2021perceiver}. The trainable readout $\bm{r}_{\omega}$ updates learned queries through self-attention and cross-attention, with $\bm{h}_k$ serving as keys and values.
The resulting tokens $\bm{c} = \bm{r}_{\omega}(\bm{h}_k)$ provide the action decoder with action-relevant predictive visual representations.

\paragraph{Action decoder}
Conditioned on the action-facing tokens $\bm{c}$, instruction $\boldsymbol{\ell}$, proprioceptive state $\bm{q}$, and optional context $\bm{s}$, the action decoder predicts the action chunk.
We instantiate the action decoder $\bm{u}_{\phi}$ with a cross-attention-conditioned video Diffusion Transformer (DiT)~\citep{peebles2022dit}.

Given an action chunk $\bm{a}_{\ast}$, the action decoder is trained with a flow matching objective: 
\begin{align}
    \mathcal{L}_{\text{act}} \coloneqq \mathbb{E}_{\bm{a}_{\ast},\bm{\epsilon},t} \Big[\big\| \bm{u}_{\phi}(\bm{a}_t,t \,| \bm{c}, \bm{q}, \bm{s}, \boldsymbol{\ell}) - (\bm{\epsilon} - \bm{a}_{\ast}) \big\|^2_2\Big] ,
\end{align}
where $\bm{\epsilon}$ is Gaussian noise, $t\in[0,1]$, and $\bm{a}_{t}=(1 - t)\bm{a}_{\ast} + t \bm{\epsilon}$.

\paragraph{Inference Without Iterative Video Denoising}
During inference, the predictive visual encoder processes the observation and instruction in a single forward pass. The resulting predictive visual representation is transformed by the readout into action-facing tokens, and the action decoder conditions on these to generate an action chunk using $N$ denoising steps. This enables real-time closed-loop control up to 11Hz even for a 14B video backbone.

%% file: 5_experiments.tex
\section{Experiments}

We validate the performance of \metabbr in both simulation and real-world environments. In particular, we investigate the following questions:
\begin{enumerate}[label=\textcolor{custom_navy}{(\textbf{Q\arabic*})}, itemsep=0.5mm, parsep=2pt, leftmargin=30pt]
    \item \label{Q1}Does distillation make visual features more useful for action prediction?
    \item \label{Q2}Do representations learned by \metabbr translate into strong manipulation performance?
    \item \label{Q3}Can \metabbr translate stronger video priors and additional video data into better control?
    \item \label{Q4}How effectively and efficiently does \metabbr perform real-world manipulation?
    \item \label{Q5}Can \metabbr learn new tasks from video demonstrations without action annotations?
\end{enumerate}

\subsection{Experimental setup}

\paragraph{Implementation details}
We use pretrained Wan2.2-5B and Wan2.2-A14B high-noise expert~\citep{wan2025wan} as the video backbone, including the video DiT, text encoder, and video VAE. The action decoder is a 1B DiT. For all experiments, we set the action horizon to 32. Following Fast-WAM~\citep{yuan2026fastwam}, we temporally downsample the videos by 4$\times$, producing 9 video frames per chunk. \metabbr is trained on each dataset without cross-embodiment pretraining. We provide more details in Appendix~\ref{app:details}.

\paragraph{Benchmarks}
We evaluate our method on two simulation benchmarks and one real-world evaluation setup. For \textbf{LIBERO}~\citep{liu2023libero}, we follow the standard benchmarking protocol, using four task suites (Spatial, Object, Goal, Long). For \textbf{RoboCasa-Kitchen}~\citep{nasiriany2024robocasa}, we follow the 24-task benchmark protocol of \citet{kim2026cosmospolicy} and train on their released dataset, which contains 50 demos per task. The tasks span object placement and manipulation of kitchen fixtures and appliances. For \textbf{real-world evaluation}, we use the DROID~\citep{khazatsky2024droid} setup, evaluating on 10 tasks selected from the pretraining distribution spanning diverse capabilities, including push, cover, press, fold, pour, and composite skills (see Appendix~\ref{app:droid_eval}). We measure the task progress and success rate using 5 rollouts per task. We provide task descriptions and evaluation details in Appendix~\ref{app:droid_eval}.

\paragraph{Baselines}
On simulation benchmarks, we compare \metabbr against state-of-the-art VLAs~\citep{rss2025_pi0,intelligence2025pi_05,kim2025openvla-oft,bjorck2025gr00t,zheng2025flare} and WAMs~\citep{bi2026motus,li2026lingbotva,yuan2026fastwam,kim2026cosmospolicy}.
For real-world DROID evaluation, we compare with state-of-the-art VLA $\pi_{0.5}$ and WAM DreamZero~\citep{ye2026dreamzero} with their publicly released checkpoints, namely $\pi_{0.5}$-DROID and DreamZero-DROID, following their inference setup.

\begin{figure}[t!]
    \centering
    \includegraphics[width=1\linewidth]{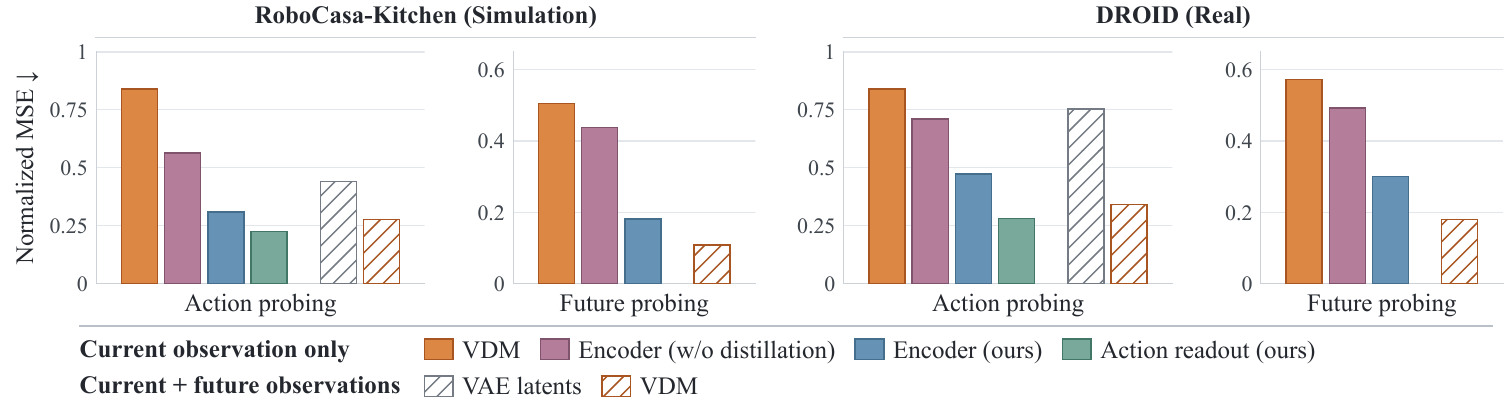}
    \captionsetup{position=bottom,skip=4pt}
    \caption{
    \textbf{Representation analysis}. We evaluate visual representations by measuring how accurately action sequences (Action probing) and future-frame VAE latents (Future probing) can be decoded from them. Future representation distillation makes the encoder's features more useful for action prediction and improves their ability to capture future visual information from the current observation.
    }
    \label{fig:rep_analysis}
\vspace{-10pt}
\end{figure}

\subsection{\texorpdfstring
{\hyperref[Q1]{\textbf{\textcolor{custom_navy}{Q1}}}}
{Q1}: Representation analysis}
\label{exp:rep_analysis}

We examine whether future representation distillation makes the visual encoder's representation more useful for action prediction. We compare features from three models given only the current observation: the pretrained video diffusion model (VDM), a variant of the \metabbr visual encoder trained without distillation, and the visual encoder of \metabbr. As a reference with direct access to future observations, we also evaluate VDM representations computed from both the current frame and ground-truth future frames.
The VDM and encoders use the 14B video backbone and are conditioned on the language instructions. Each encoder is trained separately on Robocasa-Kitchen and DROID datasets.
For each representation and input setting, we train a separate probe while keeping the feature extractor frozen. We report normalized mean squared error in Figure~\ref{fig:rep_analysis}. 

\paragraph{Action probing}
To assess the usefulness of the features for control, we train a lightweight attention-based probe to predict action sequences from each representation, using paired robot demonstration videos and actions. As an additional reference, we also probe VAE latents of both the current and ground-truth future frames to compare the encoder representations with direct encodings of future observations. Figure~\ref{fig:rep_analysis} shows that the \metabbr encoder achieves lower action prediction error than the VDM using only the current observation and the encoder variant trained without distillation. These results show that distillation makes the encoder's features more useful for action prediction.

Despite using only the current observation, \metabbr encoder achieves lower action prediction error than VAE latents of current and future frames, and approaches the accuracy of the VDM using the future frames. This indicates that the \metabbr encoder learns features that support action prediction nearly as accurately as directly observing the future. We additionally probe the features produced by the action readout module of \metabbr. These features yield even lower prediction error, demonstrating that the readout effectively extracts action-relevant information from the visual representations.

\paragraph{Future probing}
To better understand the improvement in action decodability from distillation, we examine whether distillation improves the encoder's ability to capture future visual information from the current observation. We train a spatially shared linear probe to reconstruct the VAE latent patches of ground-truth future frames. Figure~\ref{fig:rep_analysis} shows that \metabbr encoder achieves lower reconstruction error than the VDM using only the current observation and the encoder variant trained without distillation. Furthermore, its reconstruction error approaches that of the VDM given future frames on RoboCasa-Kitchen, while a gap remains on DROID. These results show that our encoder effectively captures future visual information from the current observation.



\begin{figure}
\centering
\begin{minipage}{0.45\linewidth}
    \centering
\captionsetup{position=top,skip=3pt}
\captionof{table}{
    \textbf{Average success rate (\%) across the four LIBERO task suites.}
}\label{tab:libero}
    \input{table/libero}
\end{minipage}
\hfill
\begin{minipage}{0.5\linewidth}
    \centering
\captionsetup{position=top,skip=3pt}
\captionof{table}{
    \textbf{Average success rate (\%) across 24 RoboCasa-Kitchen tasks.}
}\label{tab:robocasa}
    \input{table/robocasa-kitchen}
\end{minipage}
\end{figure}

\subsection{\texorpdfstring
{\hyperref[Q2]{\textbf{\textcolor{custom_navy}{Q2}}} \&
 \hyperref[Q3]{\textbf{\textcolor{custom_navy}{Q3}}}}
{Q2 \& Q3}: Performance and scaling on simulation benchmarks}

We first assess in-distribution task execution of \metabbr in the LIBERO benchmark compared to VLAs and WAMs. In Table~\ref{tab:libero}, \metabbr achieves performance comparable to state-of-the-art baselines even without cross-embodiment pretraining, supporting the effectiveness of \metabbr's predictive visual representations for in-distribution manipulation. 

We next evaluate \metabbr on the RoboCasa-Kitchen benchmark to assess multi-task manipulation under object and scene variation, with limited demonstration data. Table~\ref{tab:robocasa} shows that \metabbr with a 14B video backbone achieves the highest average success rate of 71.7\%, demonstrating that our predictive visual representations can translate pretrained video knowledge into strong manipulation performance without generating future videos. 
Fast-WAM also predicts actions without generating future videos at inference, but underperforms the other baselines with only 44.4\% success, suggesting that using video priors without future video generation remains challenging in this setting.

Moreover, we investigate whether \metabbr benefits from a larger video backbone and more pretraining video data. Fast-WAM shows little improvement when its backbone is scaled from 5B to 14B, suggesting limited benefit from stronger video priors. In contrast, scaling our video backbone from 5B to 14B improves success rate by 13.2\,pp, with particularly large gains on pick-and-place tasks. These gains suggest that stronger video priors provide more useful representations for coordinating grasping, transport, and placement. We also use videos from the atomic-task subset of RoboCasa365~\citep{nasiriany2026robocasa} showing robot demonstrations in simulated kitchen environments. Training the visual encoder with these videos, without their action labels, further improves success. Together, these results show that \metabbr can translate stronger video priors and additional video data into better robot control, without additional action annotations.

\begin{wrapfigure}[8]{r}{0.3\columnwidth}
  \centering
\captionof{table}{\textbf{Ablation of visual encoder training objectives}.}\label{tab:obj_abl}
\vspace{-0.1in}
  \input{table/ablation}
\end{wrapfigure}
To assess how each visual encoder training objective contributes to performance, we ablate the generative objective $\mathcal{L}_{\text{gen}}$ and distillation objective $\mathcal{L}_{\text{distill}}$. As shown in Table~\ref{tab:obj_abl}, removing the distillation objective $\mathcal{L}_{\text{distill}}$ leads to a significant drop in the success rate for both datasets. Removing the generative objective $\mathcal{L}_{\text{gen}}$ also degrades the performance, although the drops are smaller compared to removing the distillation objective. These results indicate that both objectives contribute to learning predictive visual representations, with distillation playing a particularly important role in translating pretrained video knowledge into strong manipulation performance.
This is consistent with the representation analysis in Section~\ref{exp:rep_analysis}, where removing distillation reduces both action and future decodability.

\begin{figure}[t!]
    \centering
    \captionsetup{position=bottom,skip=3pt}
\includegraphics[width=1\linewidth]{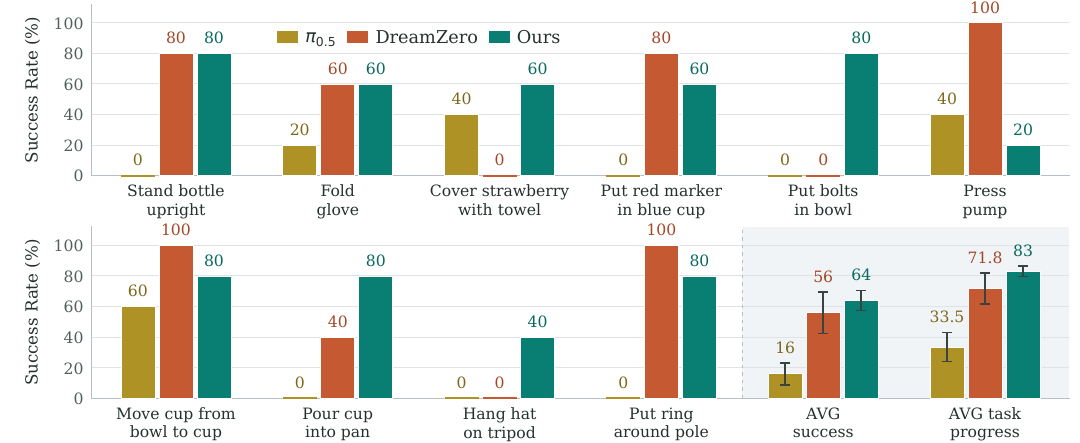}
    \caption{
    \textbf{Real-world evaluation on DROID}. We evaluate policies out of the box on 10 tasks with five trials per task. Task descriptions and the progress scoring criteria are provided in Appendix~\ref{app:droid_eval}.
    }\label{fig:droid}
\vspace{8pt}
\includegraphics[width=1\linewidth]{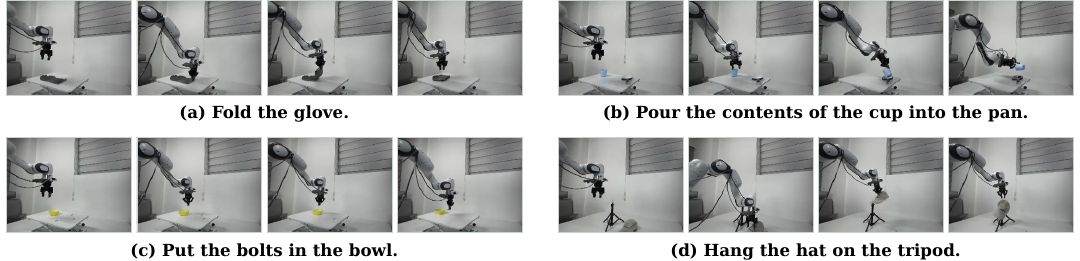}
    \caption{
    \textbf{Examples of real-world task executions}. We show selected frames from successful rollouts of \metabbr on four DROID evaluation tasks. 
    }\label{fig:droid_examples}
\end{figure}

\subsection{\texorpdfstring
{\hyperref[Q4]{\textbf{\textcolor{custom_navy}{Q4}}}}
{Q4}: Real-world performance and efficiency}

\paragraph{Real-world manipulation} To assess real-world manipulation and generalization, we evaluate the models out of the box on 10 task types represented in the pretraining data, using new objects or scene arrangements. Figure~\ref{fig:droid} reports the task progress and success rates. \metabbr achieves the highest average success rate at 64\%, compared with 56\% for DreamZero and 16\% for $\pi_{0.5}$, with successful executions spanning a broader range of task types than either baseline. 
\metabbr also achieves the highest average task progress at 83\%, compared with 71.8\% for DreamZero and 33.5\% for $\pi_{0.5}$.

The gains over DreamZero are notable in tasks involving deformable object manipulation or successive rearrangements, such as cover strawberry with towel, hang hat on tripod, and put bolts in bowl. DreamZero fails in all trials for each of these tasks, with inaccurate video predictions observed in most failures. Qualitatively, DreamZero often persists with unsuccessful actions after an initial error, whereas \metabbr often adjusts its actions and recovers. Together with the analysis in Section~\ref{exp:rep_analysis}, these observations suggest that \metabbr benefits from capturing predictive information directly in its visual representations, enabling successful control without relying on potentially inaccurate future video predictions. These results demonstrate that \metabbr achieves effective real-world control and generalizes zero-shot to new objects or scene arrangements within seen tasks. We describe the failure cases of \metabbr in Appendix~\ref{app:failure}.

\begin{figure}[t!]
    \centering
    \captionsetup{position=bottom,skip=3pt}
\includegraphics[width=1\linewidth]{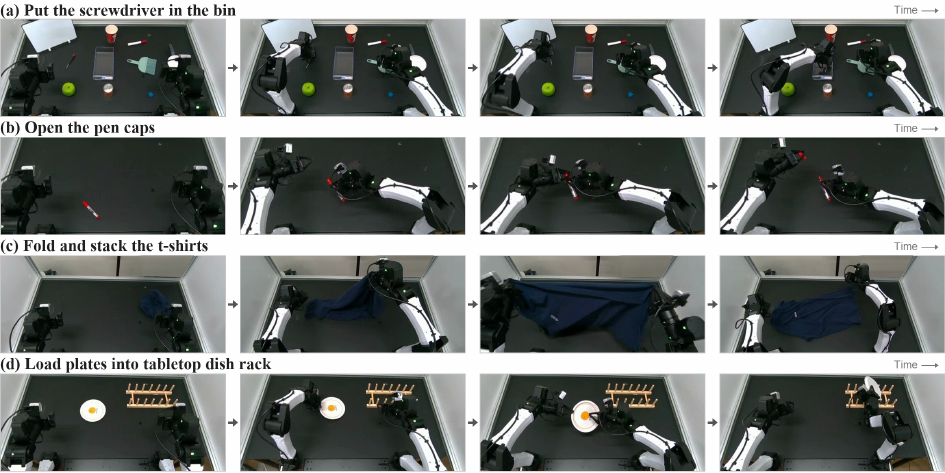}
    \caption{
    \textbf{Examples of real-world bimanual manipulation task executions}. We show selected frames from successful rollouts of \metabbr. 
    }\label{fig:yam_examples}
\vspace{-5pt}
\end{figure}

\paragraph{Real-world bimanual manipulation}
We further examine the applicability of \metabbr to another real-world embodiment using a bimanual YAM setup. We train \metabbr on a filtered 100-hour subset of ABC-130k~\citep{allshire2026abc} without additional cross-embodiment training. We deploy the resulting policy out of the box without task adaptation, evaluating tasks represented in the training set but using new scenes, objects, and arrangements.
Figure~\ref{fig:yam_examples} presents successful executions of bimanual manipulation tasks, including pick-and-place, pen uncapping, cloth folding, and dish-rack loading. These qualitative results provide evidence of zero-shot generalization to new scenes and illustrate that our approach can support precise, coordinated two-arm manipulation on a physical robot.

\paragraph{Latency}
\begin{wrapfigure}[12]{r}{0.55\columnwidth}
\centering
\vspace{-0.15in}
\includegraphics[width=1.0\linewidth]{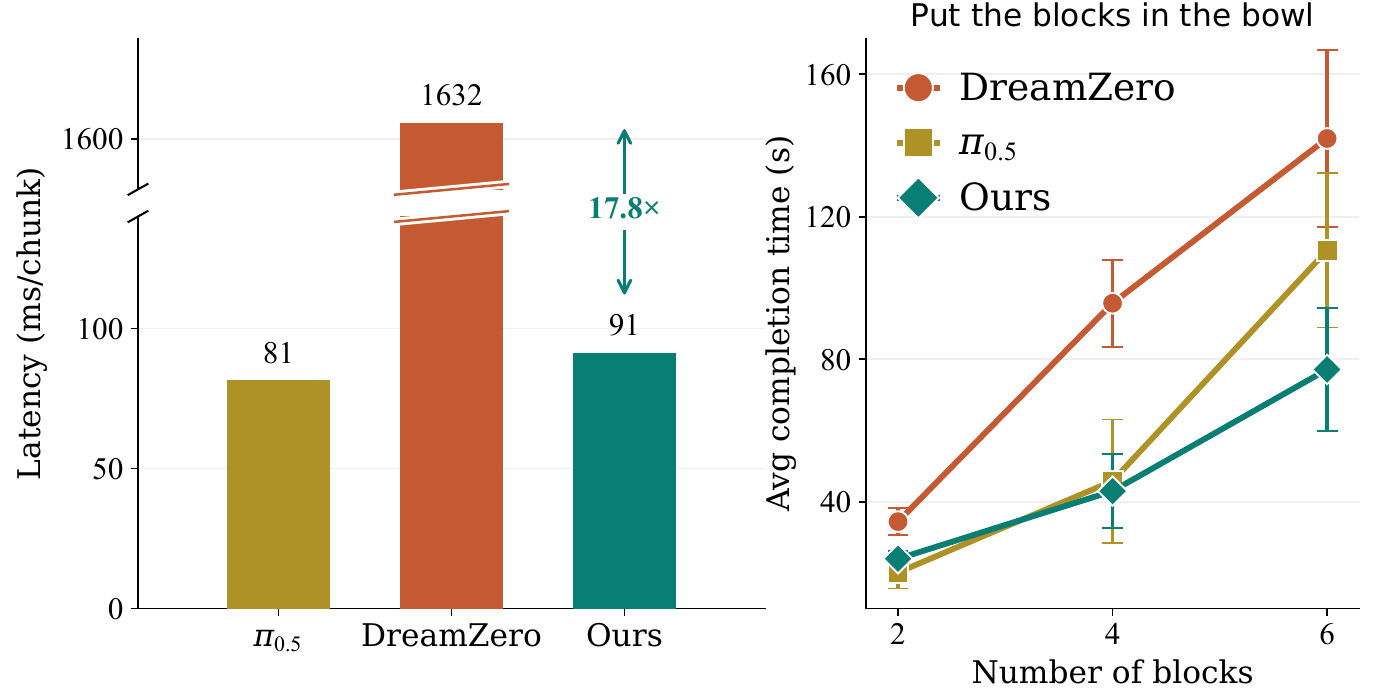}
\captionsetup{position=bottom,skip=2pt}
\caption{\textbf{Latency and completion time comparison}.}\label{fig:latency}
\end{wrapfigure}
In Figure~\ref{fig:latency} (left), we measure the mean inference latency per action chunk on NVIDIA H200 GPUs. Each policy is measured in its deployed configuration. \metabbr performs real-time closed-loop control at 11Hz, yielding a 17.8$\times$ speedup over DreamZero while achieving a higher average success rate on the real-world tasks. Compared with $\pi_{0.5}$, \metabbr achieves $4\times$ the average success rate with $1.1\times$ the latency. 

Figure~\ref{fig:latency} (right) further compares the task completion times for placing different numbers of blocks into a bowl. \metabbr achieves completion times comparable to those of $\pi_{0.5}$ on simpler settings (2 or 4 blocks), while completing the task much faster in the harder setting (6 blocks). These results highlight the efficiency of using predictive visual representations for action generation without explicit video generation.

\begin{figure}[t!]
\centering
\captionsetup{position=bottom,skip=4pt}
    \includegraphics[width=1.0\linewidth]{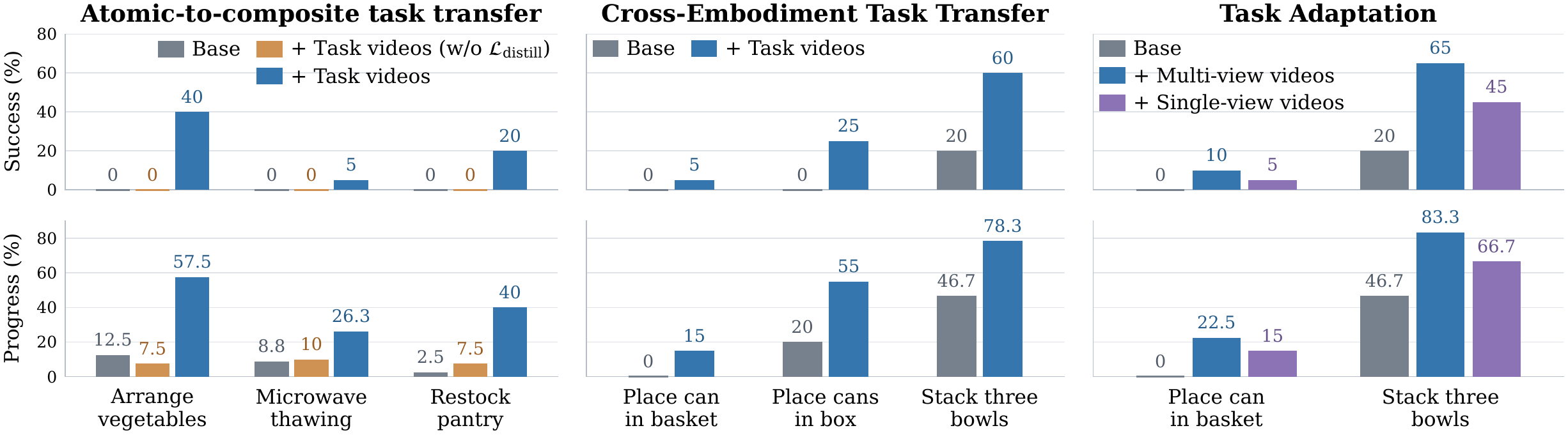}
    \caption{
    \textbf{Task transfer from video demonstrations}. We report success rates and progress over 20 episodes per task. Task descriptions are provided in Appendix~\ref{app:task_transfer}.
    }
    \label{fig:task_transfer}
\vspace{10pt}
    \includegraphics[width=1.0\linewidth]{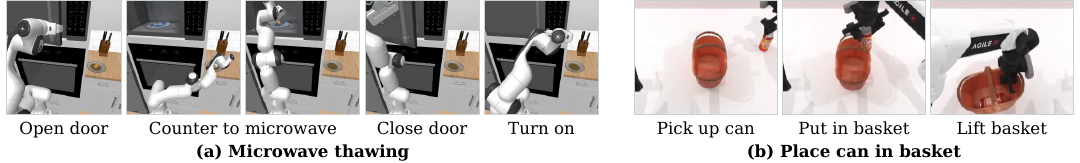}
    \caption{
    \textbf{Examples of task execution after transfer from videos}. We show selected frames from successful rollouts of \metabbr. 
    }\label{fig:task_transfer_examples}
\end{figure}

\subsection{\texorpdfstring
{\hyperref[Q5]{\textbf{\textcolor{custom_navy}{Q5}}}}
{Q5}: Task transfer from video demonstrations
}

We investigate transfer to new tasks using video demonstrations without action annotations. We consider two settings that differ in when target task videos are introduced. In the first setting, the target task videos are introduced during visual encoder pretraining, and we transfer from videos of either the same embodiment or a different embodiment. In the second, the target task videos are excluded from initial policy training and introduced for fine-tuning the policy. In both settings, we use only videos of the new tasks without action annotations.

\paragraph{Atomic-to-composite task transfer from video pretraining}
We first investigate whether \metabbr can learn new composite tasks from videos, using action labels only from atomic tasks. Atomic tasks involve individual skills while composite tasks combine several of these skills sequentially. We compare a base \metabbr trained on atomic tasks in RoboCasa-Kitchen with two variants that additionally use videos of composite tasks for visual encoder training. One variant uses only $\mathcal{L}_{\text{gen}}$ without distillation, while the other uses both $\mathcal{L}_{\text{gen}}$ and $\mathcal{L}_{\text{distill}}$. All models use the same action labels from atomic tasks, and the composite tasks are absent from the action-labeled training data.

As shown in Figure~\ref{fig:task_transfer} (left), the base model fails on all tasks, and adding the task videos with distillation raises the average success from 0\% to 21.7\%, with at least one successful trial on each task. Using the same videos without distillation yields 0\% average success. These results demonstrate that \metabbr can acquire new task knowledge from videos and transfer it to action execution without additional action annotations, with distillation playing an important role in enabling this transfer. Figure~\ref{fig:task_transfer_examples}(a) illustrates a successful composite task rollout by \metabbr, completing four subgoals in sequence.

\paragraph{Cross-embodiment task transfer from video pretraining}
We next investigate whether \metabbr can learn new tasks with task videos from a different embodiment. We use RoboTwin~\citep{chen2026robotwin} with ALOHA-AgileX as the target robot, and compare a base \metabbr trained on its atomic tasks with a variant whose encoder also learns from composite task videos recorded with a dual-arm Franka Panda setup. Both models use the same ALOHA-AgileX action labels, and the composite tasks are absent from the action-labeled training.

Figure~\ref{fig:task_transfer} (middle) shows that the base model succeeds only in bowl stacking, with 6.7\% average success across the new tasks, and adding task videos from a different embodiment improves the average success to 30\%. This demonstrates that \metabbr can transfer task knowledge from different embodiments' videos to action execution. Figure~\ref{fig:task_transfer_examples}(b) illustrates a successful task rollout by \metabbr after learning from videos of a different embodiment, without access to action-labeled demonstrations of that task.

\paragraph{Task adaptation after policy training}
Finally, we investigate whether target task knowledge can be added after policy training. We start from \metabbr trained on atomic tasks in RoboTwin, with target tasks excluded from the training set. We then fine-tune only the visual encoder separately for each task using 10 video demonstrations, keeping the action module frozen. Adaptation takes approximately 10 minutes per task on a single H200 GPU. We compare demonstrations containing multiple camera views with demonstrations containing only the head camera view.

Figure~\ref{fig:task_transfer} (right) shows that both single-view and multi-view video fine-tuning improve success rates on both tasks, with larger gains from multi-view videos. This shows that task knowledge acquired through video-only fine-tuning of the encoder can transfer to action execution without updating the action module, even with single-camera demonstrations. Figure~\ref{fig:task_transfer_examples} illustrates a successful task rollout with the fine-tuned policy.

We find that not all tasks can be learned through video demonstrations. For example, a policy unable to open cabinets also failed to learn to place an object inside a cabinet through video-only fine-tuning. This suggests that learning new tasks from videos may be limited by the policy's ability to execute the required atomic skills. We discuss these failure cases in Appendix~\ref{app:failure}.

%% file: table/libero.tex
\renewcommand{\arraystretch}{1.07}
\renewcommand{\tabcolsep}{16pt}
\small
\begin{tabular}{l c}
\toprule
     Method & Avg. SR \\
\midrule
    \textit{VLAs} & \\
    $\pi_0$~\citep{rss2025_pi0} & 94.1 \\
    $\pi_{0.5}$~\citep{intelligence2025pi_05} & 96.9 \\
    OpenVLA-OFT~\citep{kim2025openvla-oft} & 97.1 \\
\midrule
    \textit{WAMs} & \\
    Motus~\citep{bi2026motus} & 97.7 \\
    LingBot-VA~\citep{li2026lingbotva} & 98.5 \\
    Fast-WAM~\citep{yuan2026fastwam} & 97.6 \\
\midrule
    \metabbr (Ours) & \textbf{98.6} \\
\bottomrule
\end{tabular}

%% file: table/robocasa-kitchen.tex
\renewcommand{\arraystretch}{0.95}
\renewcommand{\tabcolsep}{9pt}
    \small
\begin{tabular}{l c}
\toprule
     Method & Avg. SR \\
\midrule
    $\pi_0$~\citep{rss2025_pi0} & 62.5 \\
    GR00T-N1.5~\citep{bjorck2025gr00t} & 64.1 \\
    FLARE~\citep{zheng2025flare} & 66.4 \\
    Cosmos Policy~\citep{kim2026cosmospolicy} & 67.1 \\
    Fast-WAM~\citep{yuan2026fastwam} & 44.4 \\
    Fast-WAM 14B~\citep{yuan2026fastwam} & 44.7 \\
\midrule
     \rowcolor{gg} \metabbr (Ours) & \\
     5B video backbone & 58.5 \\
     14B video backbone & \textbf{71.7} \\
     5B video backbone + extra videos & 68.8 \\
     14B video backbone + extra videos & \textbf{74.4} \\
\bottomrule
\end{tabular}

%% file: table/ablation.tex
\renewcommand{\arraystretch}{1.0}
\renewcommand{\tabcolsep}{2pt}
\small
\begin{tabular}{l c c}
\toprule
     Method & LIBERO & RoboCasa \\
\midrule
     Ours & 98.6 & 71.7 \\
     w/o $\mathcal{L}_{\text{gen}}$ & 98.0 & 55.6 \\
     w/o $\mathcal{L}_{\text{distill}}$ & 94.6 & 47.2 \\
\bottomrule
\end{tabular}

%% file: 6_conclusion.tex
\section{Conclusion}

We presented \metabbr, a WAM that repurposes pretrained video diffusion models as predictive visual encoders through future representation distillation. By distilling video priors from the intermediate representations of a frozen video diffusion model with access to future frames, \metabbr learns to predict future visual representations that guide action prediction without generating future videos. 
Experiments demonstrate strong manipulation performance in both simulation and the real world, with a significant inference speedup over video-generating WAM. \metabbr benefits from a larger video backbone and more video data, and enables task transfer from video demonstrations. These findings highlight predictive representation learning as an effective way to translate pretrained video knowledge into robot control and expand robot capabilities through video-based learning.

\section*{Acknowledgments}
This work was supported by Institute for Information \& communications Technology Planning \& Evaluation (IITP) grant funded by the Korea government (MSIT) (RS-2019-II190075, Artificial Intelligence Graduate School Program (KAIST)). National Research Foundation of Korea (NRF) grant funded by the Korea government (MSIT) (No. RS-2023-00256259), and the “Advanced GPU Utilization Support Program” funded by the Government of the Republic of Korea (MSIT).
We thank Shusheng Yang and NYU VisionX members for providing the paper template.

%% file: 7_appendix.tex
\appendix
\onecolumn

\vspace{0.5in}
\begin{center}{\bf {\LARGE Appendix}}\end{center}

\section{Model architecture details}\label{app:details}

\subsection{Video backbone adaptation}
We fine-tune the pretrained video diffusion backbone~\citep{wan2025wan} on robot videos to adapt it to the robotics domain and our multi-view packing format. Due to computational constraints, we do not use additional external video data in this stage. The adapted checkpoint initializes both the teacher and student of the same architecture. During visual-encoder training, the teacher remains frozen with bidirectional video attention and text cross-attention on all frames. The student uses the modifications below. The VAE and text encoder remain frozen throughout adaptation and visual-encoder training.

\subsection{Visual encoder}

\paragraph{Video DiT}
Following Fast-WAM~\citep{yuan2026fastwam}, we prevent current observation tokens from attending to future latent tokens. To encourage the current observation representation to capture planning-related information, we restrict each future latent frame to attend only to the current observation frame and itself, and disable text cross-attention for future latent tokens. Spatial attention is preserved within each frame.

Unlike Wan2.2-TI2V-5B, Wan2.2-I2V-A14B uses a separate image-conditioning mechanism, where a sequence consisting of the first image followed by zero-valued frames is VAE-encoded. To adopt the same first-frame conditioning strategy as TI2V-5B, we replace the first frame of the noisy video-latent stream with its clean latent and assign it timestep zero.

\paragraph{Projector}
To match the teacher's spatiotemporal token sequence length, the projector expands $N$ current observation tokens to $FN$ tokens, where $F$ is the number of latent frames in the teacher’s future representation. The student and teacher use the same spatial token grid. The student tokens are replicated $F$ times and augmented with learned frame embeddings, then fed into a two-layer decoder as queries. Each decoder layer applies self-attention, cross-attention to the original student tokens, and a $1024\rightarrow4096\rightarrow1024$ GELU MLP, with pre-normalization and residual connections in each sublayer.

Student and teacher features are extracted after complete Transformer blocks $i$ and $j$, respectively. To supervise future positions only, we exclude current observation tokens from both projector outputs and teacher targets. The loss is the mean cosine distance over the remaining token pairs after non-affine feature-wise LayerNorm ($\epsilon=10^{-6}$).

\subsection{Action readout module}
During action-module training, the visual encoder is frozen. The readout and conditioning layers are trained from scratch. The readout receives current observation tokens from encoder block $k$, independently selected from distillation block $i$, without passing through the alignment projector. 
To obtain a fixed-length visual context for the action decoder, we use 64 learned queries updated by two blocks. Each decoder layer applies self-attention, cross-attention to the projected visual tokens, and a $1024\rightarrow4096\rightarrow1024$ GELU MLP. All sublayers use pre-normalization and residual connections. Output LayerNorm is followed by a separate $1024\rightarrow1024$ conditioning projection and affine LayerNorm. These visual tokens are appended to the projected instruction tokens as decoder cross-attention context.

\subsection{State and action encoders}
Separate linear projections map the normalized proprioceptive state and each noisy action to the transformer's embedding dimension. The state contributes one prefix token, replaced with a learned mask token with probability 0.2 during training. 

\subsection{Action decoder}
The decoder uses 30 DiT blocks, hidden width 1024, feed-forward width 4096, and 24 attention heads of dimension 128, totaling approximately 1B parameters. Each block contains bidirectional self-attention over state/history/action tokens and cross-attention to instruction and visual context from the action readout module. Self-attention uses one-dimensional rotary position embeddings and padding masks. A 256-dimensional sinusoidal timestep embedding and an MLP provide block modulation. A final linear layer predicts the flow-matching velocity in action space, and prefix outputs are discarded, and loss is computed only at valid future-action positions.

For DROID, we include a history of the previous 8 actions to address unintended gripper reopening during closed-loop execution. History uses the same linear encoder as noisy actions. The input order is current state, action history, then noisy future actions. Missing history entries are masked, and the entire history is independently dropped with probability 0.2 during training. State and history remain noise-free, with a fixed timestep corresponding to $\sigma=10^{-3}$.

\subsection{Training and inference algorithms}\label{app:algorithms}
Algorithms~\ref{alg:dreamless_train} and~\ref{alg:dreamless_inference} summarize training and synchronous closed-loop inference. We write $\bm{x}_{\ast}$ for VAE-encoded video latents, with superscripts $0$ and $+$ selecting current observation and future positions. Video-model calls also include the Wan-A14B image-conditioning channels when applicable. $\operatorname{sg}$ denotes stop-gradient, $\operatorname{LN}$ is non-affine token-wise LayerNorm, and $\mathrm{sim}$ is mean cosine similarity after this normalization. States and actions are normalized. $\bm{s}$ denotes optional action history, used only for DROID.

Here, $p_{\mathrm{train}}$ samples $t=5r/(1+4r)$ with $r\sim\mathcal{U}(0,1)$, and $w(t)$ is the native scheduler's timestep weight. Losses are averaged over the batch. $\operatorname{MSE}_{m}$ excludes padded action positions. The implemented cosine loss $1-\mathrm{sim}$ differs from $-\mathrm{sim}$ only by an additive constant. Teacher conditioning keeps the first frame clean for 5B, and Wan-A14B retains a noised first frame in the video stream and separate clean image conditioning.

\begingroup
\captionsetup{hypcap=false}
\begin{figure}[t]
\centering
\begin{minipage}[t]{0.51\textwidth}
\vspace{0pt}
\input{algorithm/training}
\end{minipage}\hfill
\begin{minipage}[t]{0.47\textwidth}
\vspace{0pt}
\input{algorithm/inference}
\end{minipage}
\end{figure}
\endgroup

\section{Implementation details}\label{app:imp-detail}

\subsection{Baselines}
For simulation benchmarks, we use reported baseline scores except for Fast-WAM on RoboCasa-Kitchen, which we train and evaluate ourselves.

\paragraph{Fast-WAM~\citep{yuan2026fastwam}} 
For RoboCasa-Kitchen, we train Fast-WAM 5B using the official implementation with learning rate $10^{-4}$ for 20 epochs and global batch size 128. The 14B baseline uses the Wan2.2-A14B high-noise expert with global batch size 32. We modify the conditioning on the current observation frame to match that of our method.

\subsection{\metabbr}
We use AdamW with a learning rate of $10^{-4}$ for both training stages. For LIBERO and Robocasa models, we use rank-128 LoRA to train the visual encoders initialized from the video diffusion models. For the 14B DROID model, we fully train the weights without using LoRA.  Input resolutions are $224\times448$ for LIBERO, $352\times352$ for RoboCasa, and $352\times640$ for DROID. We set the teacher noise level to $t_d=0.7$ for simulation benchmarks and $0.8$ for DROID.

The layer indices used for our distillation loss and action readout are fixed for each backbone size across all datasets and tasks. For the 5B backbone, we extract student features for distillation at $i=25$ and features for action readout at $k=19$. For the 14B backbone, we use $i=j=k=31$. All indices are zero-based, and features are extracted after the corresponding complete Transformer block.

For LIBERO, we use global batch sizes of 128 and 64 in stages 1 and 2, respectively. Stage 1 runs for 40 epochs, and stage 2 for 10 epochs. For RoboCasa 5B, we use global batch sizes of 64 and 256 in stages 1 and 2, respectively, for four and 20 epochs. For RoboCasa 14B, we use global batch sizes of 32 and 512 in stages 1 and 2, respectively. Stage 1 runs for 10 epochs on RoboCasa-only videos or two epochs with additional videos, and stage 2 runs for 20 epochs. For DROID, we fully train the visual encoder in stage 1 without LoRA. We use global batch sizes of 64 and 4,096 in stages 1 and 2, respectively, for 6 and 16 epochs. Training across both stages took approximately one week on 16 NVIDIA H200 GPUs.

For ABC-130k~\citep{allshire2026abc}, we train our model on a subset of approximately 100 hours from the full 3,500-hour dataset due to computational constraints. The subset is selected to cover diverse tasks and operators. We use the 14B backbone with an input resolution of $432\times384$. For the bimanual setup, the overhead view occupies the upper two-thirds of the input, while the left and right wrist views are arranged side by side in the lower third. We follow the DROID 14B optimization settings, including global batch sizes of 64 and 4,096 for stages 1 and 2, respectively.

\paragraph{Inference}
The visual encoder runs once at timestep zero, without video denoising or classifier-free guidance (CFG). Actions are initialized from Gaussian noise and sampled with 10 Euler steps for LIBERO and RoboCasa, and 16 steps for DROID, using a flow-matching schedule with shift 5. Action sampling also omits CFG, requiring one conditional decoder evaluation per step. From each 32-action chunk, we execute 10 actions in LIBERO, 32 in RoboCasa, and 24 in DROID before replanning.

\section{Experimental details}

\subsection{Simulation benchmarks}

\paragraph{LIBERO}
We use the same multi-view frame packing setup as in the official Fast-WAM implementation. We report the average success rate across the four LIBERO suites, evaluated over a total of 2,000 trials covering 40 tasks with different random seeds. The baseline results are taken from \citet{yuan2026fastwam}.

\paragraph{Robocasa-Kitchen}
We use the Cosmos Policy~\citep{kim2026cosmospolicy} dataset with 50 human-teleoperated demonstrations per task. We train and evaluate \metabbr and Fast-WAM on this benchmark and use reported scores for the remaining baselines. Following DreamZero~\citep{ye2026dreamzero}, we pack three camera views into a single frame, with the wrist view stretched horizontally by $2\times$ above two side-by-side external views. Each of the 24 tasks is evaluated over 50 trials. The baseline results are taken from \citet{kim2026cosmospolicy}.

For our additional-video experiments, we use a filtered and converted version of the RoboCasa365 human-teleoperation atomic-task dataset. The source dataset covers 65 atomic tasks spanning ten skill families. Each atomic task involves one basic skill executed over multiple timesteps, whereas composite tasks combine several skills. Demonstrations use a Panda arm mounted on an Omron mobile base, with synchronized left, right, and wrist camera videos. We use the additional videos for backbone adaptation and visual-encoder training without action supervision. Action-module training uses only the Cosmos Policy demonstrations.

\subsection{Real-world experiments}\label{app:droid_eval}

\input{table/eval_tasks}

\paragraph{Tasks and evaluation}
We evaluate on 10 tasks from the pretrained distribution, performing 5 rollouts per task, for a total of 50 evaluation rollouts for each model. Table~\ref{tab:droid-eval-task-descriptions} describes the ten evaluation tasks, and Table~\ref{tab:droid-eval-progress-rubrics} describes their progress scores. Following \citet{barreiros2025overlay}, we apply an image overlay to the initial scene to reduce variance.

\paragraph{Latency}
Although single-pass visual encoding without iterative video denoising or CFG already makes \metabbr faster than DreamZero, we further reduce latency through engineering optimizations. These include VAE encoding of only the current frame, fused attention kernels, and fused QKV/KV projections. Visual and action Transformer blocks are compiled with \texttt{torch.compile}, and CUDA graphs replay VAE encoding, visual feature extraction, and the action sampling loop to reduce launch overhead. Instruction embeddings are cached across queries, while action-conditioning keys and values are computed once per chunk and reused across denoising steps.

We measure latency on a single NVIDIA H200 after startup warm-up, including client preprocessing, server policy execution, and client postprocessing, and excluding communication and robot execution.

Regarding task completion time, we use three rollouts per policy and block-count setting. We estimate each rollout's duration by adding nominal robot execution time to its total policy computation time. Communication and setup overhead are excluded.

\input{table/transfer_tasks}

\subsection{Task transfer from video demonstrations}\label{app:task_transfer}

We provide task descriptions in Table~\ref{tab:task-transfer-descriptions}. In the first two experiments, target task videos are included during visual encoder pretraining, before training the action module on atomic tasks. In the third experiment, target task demonstrations are excluded from initial policy training, and target task videos are used only for subsequent adaptation. In all three experiments, the target task demonstrations contain no action annotations.

\paragraph{Atomic-to-composite task transfer from video pretraining}
We include videos of three target RoboCasa composite tasks from the same embodiment during visual encoder pretraining. We merge the existing LoRA weights into the encoder, freeze the merged weights, and train fresh rank-128 adapters and the alignment projector on the target task videos at a learning rate of $10^{-4}$. After encoder pretraining, we freeze the encoder and train the action module using only action-labeled demonstrations of atomic tasks. Across all three settings, the instruction remains fixed throughout each rollout, without intermediate subgoal prompts. Progress is the fraction of subgoals achieved in order.

\paragraph{Cross-embodiment task transfer from video pretraining}
We include videos of three target composite tasks recorded with dual-arm Franka Panda during visual encoder pretraining. We use 50 video demonstrations per task, totaling 150 demonstrations. We then freeze the encoder and train the action module using only ALOHA-AgileX atomic-task demonstrations. We evaluate the three target tasks on ALOHA-AgileX, testing whether task knowledge learned from videos of a different embodiment transfers to action execution.

\paragraph{Task adaptation after policy training}
We start from a policy trained on RoboTwin atomic tasks, with target task demonstrations excluded from both visual encoder pretraining and action-module training. We then fine-tune only the visual encoder separately for each target task using 10 video demonstrations, while keeping the trained action module frozen. We compare fine-tuning with multiple camera views and with only the head camera: basket placement uses 60 and 40 iterations, respectively, and bowl stacking uses 100 and 80. Both variants are evaluated in RoboTwin using all three camera views.

\section{Additional results}

\input{table/libero_per_suite}
\input{table/robocasa_per_task}

\subsection{LIBERO per-suite results}\label{app:add_libero}
Table~\ref{tab:libero_per_suite} reports per-suite success rates for \metabbr on LIBERO.

\subsection{{Robocasa kitchen per-task results}}
Table~\ref{tab:robocasa_per_task} reports per-task success rates for the 14B model trained with additional video data.

\subsection{Failure case analysis}\label{app:failure}

\paragraph{Real-world manipulation}
Although \metabbr improves over DreamZero on tasks involving deformable objects and successive rearrangements, pump pressing remains a failure case. The policy contacts the pump head in all five trials but completes the task in only one. We hypothesize that this reflects difficulty generalizing the motion required to actuate a new object, even when the target is reached. Whether broader training coverage of such interactions improves this generalization remains a direction for future work.

\paragraph{Task transfer}
The cabinet insertion task requires opening the cabinet before placing an object inside. The base policy did not reliably perform the opening motion, and video-only fine-tuning did not enable it to open. This failure suggests that transferring task knowledge from videos may depend on the availability of related action skills in the base policy. With the action module frozen, adapting the visual encoder alone may be insufficient to acquire a new motor skill. We leave it to future work to test whether scaling pretraining with more diverse action demonstrations alleviates this limitation.

\section{Further discussions}
\subsection{Connection to prior works}
\paragraph{Comparison with Fast-WAM~\citep{yuan2026fastwam}}
Fast-WAM uses a pretrained video diffusion backbone for direct action prediction, retaining video co-training but omitting future-video denoising at deployment. The key difference is that \metabbr directly supervises the policy features with future representation distillation in addition to the video generative objective. In our RoboCasa-Kitchen experiments (Table~\ref{tab:robocasa}), Fast-WAM substantially trails the video-generating WAM baseline and shows little improvement from a 5B to a 14B backbone (44.4\% to 44.7\%). These results suggest that generative supervision through shared weights alone does not ensure that a larger VDM provides more useful predictive features for control. \metabbr supports learning and transfer from videos without action labels, while achieving performance and generalization competitive with video-generating WAMs at substantially lower inference latency.

\paragraph{Knowledge Insulation~\citep{driess2025knowledge}}
Freezing the visual encoder in stage 2 prevents gradients from the action DiT from updating the video backbone. Knowledge Insulation and LAP~\citep{zha2026lap} similarly block gradients from the action expert while training VLM representations through FAST-token prediction and language-action prediction, respectively. Our method applies a similar separation to a VDM, using the native video generative objective and future representation distillation to shape the visual features. Their concurrent training also suggests an alternative to our two-stage procedure: jointly training the encoder and action module with their respective objectives, while blocking action-module gradients from the encoder. We leave this variant to future work, as we have not evaluated it due to computational constraints.

\subsection{Limitations}
Although \metabbr uses video-model priors to predict actions directly like VLAs, achieving lower latency than video-generating WAMs, several limitations remain.

\begin{itemize}[itemsep=0.5mm, parsep=3pt, leftmargin=*]
    \item \textbf{Scaling with embodied pretraining.} We demonstrate that a 14B video backbone trained on DROID without embodied pretraining can outperform existing VLA and WAM baselines. Although the RoboCasa results in Table~\ref{tab:robocasa} suggest that additional video data could further improve performance, we did not evaluate this extension on DROID due to computational constraints. Scaling to larger backbones, datasets, and embodiments remains an important direction for future work.
    \item \textbf{Training complexity.} Our pipeline separates backbone adaptation, predictive representation learning, and action learning. Simplifying these stages and exploring joint training of the visual encoder and action module remain directions for future work.
\end{itemize}

%% file: algorithm/training.tex
\begin{algorithm}[H]
\caption{\metabbr training}
\label{alg:dreamless_train}
\small
\begin{algorithmic}[1]
\REQUIRE Data $\mathcal{D}_{v},\mathcal{D}_{a}$; fixed teacher noise $t_d$
\STATE Adapt video backbone on $\mathcal{D}_{v}$; initialize $\bm{g}$ and $\bm{u}_{\theta}$.
\STATE Freeze $\bm{g}$; initialize projector $\bm{p}_{\psi}$.
\STATE \textbf{Stage 1: Visual encoder}
\WHILE{training}
    \STATE Sample $(\bm{x}_{\ast},\boldsymbol{\ell})\sim\mathcal{D}_{v}$ and $t,\bm{\epsilon},\bm{\epsilon}_d$.
    \STATE $\bm{x}_{t}\gets(1-t)\bm{x}_{\ast}+t\bm{\epsilon}$; $\bm{x}_{t}^{0}\gets\bm{x}_{\ast}^{0}$.
    \STATE $(\hat{\bm{v}},\bm{h}_i)\gets\bm{u}_{\theta}(\bm{x}_{t},t,\boldsymbol{\ell})$. \COMMENT{$\bm{h}_i$: current frame}
    \STATE $\bm{x}_{t_d}\gets(1-t_d)\bm{x}_{\ast}+t_d\bm{\epsilon}_d$.
    \STATE $\bm{y}\gets[\bm{g}_j(\bm{x}_{t_d},t_d,\boldsymbol{\ell})]^{+}$. \COMMENT{$+$: future tokens only}
    \STATE $\mathcal{L}_{\mathrm{gen}}\gets w(t)\operatorname{MSE}(\hat{\bm{v}}^{+},(\bm{\epsilon}-\bm{x}_{\ast})^{+})$.
    \STATE $\mathcal{L}_{\mathrm{distill}}\gets-\mathrm{sim}([\bm{p}_{\psi}(\bm{h}_i)]^{+},\bm{y})$.
    \STATE Update $\theta,\psi$ with $\mathcal{L}_{\mathrm{gen}}+\mathcal{L}_{\mathrm{distill}}$.
\ENDWHILE
\STATE \textbf{Stage 2: Action module}
\STATE Freeze $\bm{u}_{\theta}$; initialize the action module.
\WHILE{training}
    \STATE Sample $(\bm{x}_{\ast}^{0},\boldsymbol{\ell},\bm{q},\bm{s},\bm{a}_{\ast})\sim\mathcal{D}_{a}$ and $t,\bm{\epsilon}$.
    \STATE $\bm{h}_k\gets[\bm{u}_{\theta}(\bm{x}_{\ast}^{0},0,\boldsymbol{\ell})]_k$.
    \STATE $\bm{c}\gets\bm{r}_{\omega}(\operatorname{LN}(\bm{h}_k))$. \COMMENT{Action readout}
    \STATE $\bm{a}_{t}\gets(1-t)\bm{a}_{\ast}+t\bm{\epsilon}$.
    \STATE $\hat{\bm{v}}_a\gets\bm{u}_{\phi}(\bm{a}_{t},t\mid\bm{c},\bm{q},\bm{s},\boldsymbol{\ell})$.
    \STATE $\mathcal{L}_{\mathrm{act}}\gets w(t)\operatorname{MSE}(\hat{\bm{v}}_a,\bm{\epsilon}-\bm{a}_{\ast})$.
    \STATE Update the action module with $\mathcal{L}_{\mathrm{act}}$.
\ENDWHILE
\end{algorithmic}
\end{algorithm}

%% file: algorithm/inference.tex
\begin{algorithm}[H]
\caption{\metabbr inference}
\label{alg:dreamless_inference}
\small
\begin{algorithmic}[1]
\REQUIRE Instruction $\boldsymbol{\ell}$, steps $N$, execution length $R$
\STATE $\{t_n\}_{n=0}^{N}\gets\operatorname{Schedule}(N)$.
\STATE Initialize context $\bm{s}\gets\varnothing$.
\WHILE{task not done}
    \STATE $(\bm{o},\bm{q})\gets\operatorname{Observe}()$; $\bm{x}_{\ast}^{0}\gets\operatorname{VAE}(\bm{o})$.
    \STATE $\bm{h}_k\gets[\bm{u}_{\theta}(\bm{x}_{\ast}^{0},0,\boldsymbol{\ell})]_k$. \COMMENT{One single pass}
    \STATE $\bm{c}\gets\bm{r}_{\omega}(\operatorname{LN}(\bm{h}_k))$.
    \STATE Sample $\bm{a}_{t_0}\sim\mathcal{N}(\bm{0},\bm{I})$.
    \FOR{$n=0,\ldots,N-1$}
        \STATE $\hat{\bm{v}}_n\gets\bm{u}_{\phi}(\bm{a}_{t_n},t_n\mid\bm{c},\bm{q},\bm{s},\boldsymbol{\ell})$.
        \STATE $\bm{a}_{t_{n+1}}\gets\bm{a}_{t_n}+(t_{n+1}-t_n)\hat{\bm{v}}_n$.
    \ENDFOR
    \STATE Postprocess $\bm{a}_{t_N}$; execute its first $R$ actions.
    \IF[DROID]{history enabled}
        \STATE Update action history in $\bm{s}$.
    \ENDIF
\ENDWHILE
\end{algorithmic}
\end{algorithm}

%% file: table/eval_tasks.tex
\begin{table*}[t]
  \centering
  \small
  \setlength{\tabcolsep}{4pt}
  \renewcommand{\arraystretch}{1.1}
  \caption{Real-world evaluation tasks on DROID.}
  \label{tab:droid-eval-task-descriptions}
  \begin{tabular}{@{} >{\raggedright\arraybackslash}p{0.20\textwidth} >{\raggedright\arraybackslash}p{0.28\textwidth} >{\raggedright\arraybackslash}p{\dimexpr0.52\textwidth-4\tabcolsep\relax} @{}}
    \toprule
    \textbf{Task} & \textbf{Instruction} & \textbf{Setup and objective} \\
    \midrule
    \textbf{Stand bottle upright} & ``Stand the bottle upright.'' & A bottle lies on its side. Stand it on its base and release it. \\
    \addlinespace[3pt]
    \textbf{Fold glove} & ``Fold the glove.'' & A glove lies unfolded. Make a regular fold that turns back more than one-third of its original length and remains after release. \\
    \addlinespace[3pt]
    \textbf{Cover strawberry with towel} & ``Use the towel to cover the strawberry.'' & A toy strawberry and grape lie on opposite sides of a towel. Lift the towel and release it over the strawberry, fully covering it while leaving the grape uncovered. \\
    \addlinespace[3pt]
    \textbf{Put red marker in blue cup} & ``Put the red marker inside the blue cup.'' & Red and black markers lie beside blue and green cups. Release the red marker inside the upright blue cup; the black marker and green cup are distractors. \\
    \addlinespace[3pt]
    \textbf{Move cup from bowl to cup} & ``Pick up the cup from the bowl and put it in the other cup.'' & One cup sits inside a bowl, with another cup and a distractor can nearby. Remove the source cup and release it inside the other cup. \\
    \addlinespace[3pt]
    \textbf{Press pump} & ``Press the pump.'' & An upright pump bottle stands beside a cup holding a toothbrush. Depress the pump head while keeping the bottle upright. \\
    \addlinespace[3pt]
    \textbf{Put bolts in bowl} & ``Put the bolts in the bowl.'' & Three bolts lie separately beside a bowl. Place and release all three inside the bowl. \\
    \addlinespace[3pt]
    \textbf{Pour cup into pan} & ``Pour the contents of the cup into the pan.'' & A source cup and receiving pan sit on the table. Transfer the cup's contents into the pan. \\
    \addlinespace[3pt]
    \textbf{Hang hat on tripod} & ``Hang the hat on the tripod.'' & A hat lies beside an upright tabletop tripod. Place and release the hat so that the tripod supports it. \\
    \addlinespace[3pt]
    \textbf{Put ring around pole} & ``Put the ring around the pole.'' & A ring lies beside an upright pole on a base. Place and release the ring with the pole passing through its hole. \\
    \bottomrule
  \end{tabular}
\end{table*}

\begin{table*}[t]
  \centering
  \small
  \setlength{\tabcolsep}{4pt}
  \renewcommand{\arraystretch}{1.1}
  \caption{Task-progress rubrics for real-world evaluation tasks on DROID.}
  \label{tab:droid-eval-progress-rubrics}
  \begin{tabular}{@{} >{\raggedright\arraybackslash}p{0.24\textwidth} >{\raggedright\arraybackslash}p{\dimexpr0.76\textwidth-2\tabcolsep\relax} @{}}
    \toprule
    \textbf{Task} & \textbf{Scoring criteria} \\
    \midrule
    \textbf{Stand bottle upright} & $0.5$: bottle axis exceeds $45^\circ$ above horizontal; upright is $90^\circ$. $1$: bottle stands on its base after release. \\
    \addlinespace[3pt]
    \textbf{Fold glove} & $0$: no fold. $0.5$: a nonzero regular fold turns back at most one-third of the original glove length, or an irregular fold occurs, regardless of folded length. $1$: a regular fold turns back more than one-third of the original length and remains after release. \\
    \addlinespace[3pt]
    \textbf{Cover strawberry with towel} & Requires lifting, then partial coverage, then full coverage. $1/3$: towel lifted. $2/3$: any partial strawberry coverage after lifting, including slight coverage. $1$: strawberry fully covered after towel release, with the grape uncovered. Coverage without first lifting does not qualify. \\
    \addlinespace[3pt]
    \textbf{Put red marker in blue cup} & $0.25$: red marker picked up. $0.5$: red marker contacts the blue cup. $0.75$: marker enters the cup, but the cup falls or tips over. $1$: marker released inside the upright blue cup. Picking up the black marker before picking up the red marker makes the entire rollout score $0$, even after later completion; incidental contact does not trigger this penalty. \\
    \addlinespace[3pt]
    \textbf{Move cup from bowl to cup} & $1/3$: correct source cup picked up. $2/3$: source cup contacts the receiving cup without successful insertion. $1$: source cup inserted and released inside the receiving cup. \\
    \addlinespace[3pt]
    \textbf{Press pump} & $0.5$: contact with the movable pump head. $1$: pump head depressed while the bottle remains upright. Spring return after release is allowed. \\
    \addlinespace[3pt]
    \textbf{Put bolts in bowl} & $1/3$ per bolt placed and released inside the bowl. Retain intermediate credit for the highest count reached; $1$ requires all three bolts to remain inside at the end. \\
    \addlinespace[3pt]
    \textbf{Pour cup into pan} & $0.5$: source cup picked up. $1$: its contents transferred into the pan. \\
    \addlinespace[3pt]
    \textbf{Hang hat on tripod} & $1/3$: hat picked up. $2/3$: hat contacts the tripod. $1$: hat remains supported by the tripod after release. \\
    \addlinespace[3pt]
    \textbf{Put ring around pole} & $0.5$: ring picked up. $1$: ring placed and released with the pole passing through its hole. \\
    \bottomrule
  \end{tabular}
\end{table*}

%% file: table/transfer_tasks.tex
\begin{table}[t]
    \centering
    \small
    \setlength{\tabcolsep}{4pt}
    \renewcommand{\arraystretch}{1.15}
    \caption{Task descriptions and ordered subgoals for task transfer from
    video demonstrations. Subgoals are evaluated in the listed order.}
    \label{tab:task-transfer-descriptions}
    \begin{tabular}{
        @{}
        >{\raggedright\arraybackslash}p{0.15\linewidth}
        >{\raggedright\arraybackslash}p{0.23\linewidth}
        >{\raggedright\arraybackslash}p{\dimexpr0.62\linewidth-4\tabcolsep\relax}
        @{}
    }
        \toprule
        \textbf{Setting} & \textbf{Task} & \textbf{Description and ordered subgoals} \\
        \midrule
        Atomic-to-composite
        & Arrange vegetables
        & Transfer the first vegetable from the sink to the counter
          $\rightarrow$ transfer the second vegetable from the sink to the counter. \\
        \addlinespace[4pt]

        Atomic-to-composite
        & Microwave thawing
        & Open the microwave door
          $\rightarrow$ transfer the object from the counter into the microwave
          $\rightarrow$ close the door
          $\rightarrow$ turn on the microwave. \\
        \addlinespace[4pt]

        Atomic-to-composite
        & Restock pantry
        & Transfer the first object from the counter into the cabinet
          $\rightarrow$ transfer the second object from the counter into the cabinet. \\
        \midrule

        Cross embodiment
        & Place can in basket
        & Place the can inside the basket
          $\rightarrow$ lift the basket.
          Can placement alone constitutes partial completion. \\
        \addlinespace[4pt]

        Cross embodiment
        & Place cans in box
        & Place the first can inside the plastic box
          $\rightarrow$ place the second can inside the box. \\
        \addlinespace[4pt]

        Cross embodiment
        & Stack three bowls
        & Establish the base bowl
          $\rightarrow$ stack the second bowl on the base
          $\rightarrow$ stack the third bowl on the stack. \\
        \bottomrule
    \end{tabular}
\end{table}

%% file: table/libero_per_suite.tex
\begin{table}[t]
\centering
\small
\setlength{\tabcolsep}{8pt}
\caption{LIBERO success rates (\%) of \metabbr across the four suites. 
}
\label{tab:libero_per_suite}
\begin{tabular}{lccccc}
\toprule
Method & Spatial & Object & Goal & Long & Average \\
\midrule
LingBot-VA~\citep{li2026lingbotva} & 98.5 & 99.6 & 97.2 & 98.5 & 98.5 \\
Fast-WAM~\citep{yuan2026fastwam} & 98.2 & 100.0 & 97.0 & 95.2 & 97.6 \\
\midrule
\metabbr (Ours) & 99.0 & 99.8 & 98.8 & 96.6 & \textbf{98.6} \\
\bottomrule
\end{tabular}
\end{table}

%% file: table/robocasa_per_task.tex
\begin{table}[t]
\centering
\small
\setlength{\tabcolsep}{8pt}
\caption{RoboCasa-Kitchen success rates (\%) for \metabbr with the 14B backbone and additional video data. Each task is evaluated over 50 trials.}
\label{tab:robocasa_per_task}
\begin{tabular}{lrlr}
\toprule
Task & SR (\%) & Task & SR (\%) \\
\midrule
\texttt{CloseDoubleDoor} & 84 & \texttt{PnPCounterToSink} & 72 \\
\texttt{CloseDrawer} & 98 & \texttt{PnPCounterToStove} & 74 \\
\texttt{CloseSingleDoor} & 98 & \texttt{PnPMicrowaveToCounter} & 58 \\
\texttt{CoffeePressButton} & 96 & \texttt{PnPSinkToCounter} & 76 \\
\texttt{CoffeeServeMug} & 72 & \texttt{PnPStoveToCounter} & 72 \\
\texttt{CoffeeSetupMug} & 38 & \texttt{TurnOffMicrowave} & 90 \\
\texttt{OpenDoubleDoor} & 96 & \texttt{TurnOffSinkFaucet} & 86 \\
\texttt{OpenDrawer} & 88 & \texttt{TurnOffStove} & 18 \\
\texttt{OpenSingleDoor} & 90 & \texttt{TurnOnMicrowave} & 94 \\
\texttt{PnPCabToCounter} & 56 & \texttt{TurnOnSinkFaucet} & 76 \\
\texttt{PnPCounterToCab} & 62 & \texttt{TurnOnStove} & 52 \\
\texttt{PnPCounterToMicrowave} & 50 & \texttt{TurnSinkSpout} & 90 \\
\midrule
\multicolumn{3}{l}{\textbf{Mean over 24 tasks}} & \textbf{74.4} \\
\bottomrule
\end{tabular}
\end{table}